# Inappropriate Benefits and Identification of ChatGPT Misuse in Programming Tests: A Controlled Experiment

Hapnes Toba[1*], Oscar Karnalim[2], Meliana Christianti Johan[3], Terutoshi Tada[4], Yenni Merlin Djajalaksana[5], and Tristan Vivaldy[6]

[1,2] Master of Computer Science Study Program, Faculty of Information Technology, Maranatha Christian University, Bandung, Indonesia
[1]hapnestoba@it.maranatha.edu
[2]oscar.karnalim@it.maranatha.edu
[3,6]Bachelor of Informatics Study Program, Faculty of Information Technology, Maranatha Christian University, Bandung, Indonesia
[3]meliana.christianti@it.maranatha.edu
[6]2172032@maranatha.ac.id
[4]Media Culture Department, Faculty of Information Sciences and Arts, Toyo University, Kawagoe, Japan
[4]t_tada@toyo.jp
[5]EC-Council, Tampa, Florida, The United States of America
[5]yenni.djajalaksana@eccouncil.org

**Abstract.** While ChatGPT may help students to learn to program, it can be misused to do plagiarism, a breach of academic integrity. Students can ask ChatGPT to complete a programming task, generating a solution from other people's work without proper acknowledgment of the source(s). To help address this new kind of plagiarism, we performed a controlled experiment measuring the inappropriate benefits of using ChatGPT in terms of completion time and programming performance. We also reported how to manually identify programs aided with ChatGPT (via student behavior while using ChatGPT) and student perspective of ChatGPT (via a survey). Seventeen students participated in the experiment. They were asked to complete two programming tests. They were divided into two groups per the test: one group should complete the test without help while the other group should complete it with ChatGPT. Our study shows that students with ChatGPT complete programming tests two times faster than those without ChatGPT, though their programming performance is comparable. The generated code is highly efficient and uses complex data structures like lists and dictionaries. Based on the survey results, ChatGPT is recommended to be used as an assistant to complete programming tasks and other general assignments. ChatGPT will be beneficial as a reference as other search engines do. Logical and critical thinking are needed to validate the result presented by ChatGPT.

# 1 Introduction

In programming education, plagiarism is a common breach of academic integrity [1]. It is about the reuse of one's program (or even part of it) with insufficient acknowledgment to the owner [2]. While code reuse is somewhat encouraged in programming, there is a need to cite the source in the program comments [3].

Plagiarism can happen due to either pressure, opportunity, or misrationalization [4]. Students are tempted to plagiarize if they are pressured by the circumstances, they misrationalise the act with incorrect justifications, and there are opportunities to cheat (e.g., limited plagiarism detection mechanism). To deal with pressure, instructors can promote early submissions via incentives [5] or replace large assessments with many small assessments [6]. To deal with misrationalization, instructors are expected to inform students about the matter in their courses [7]. Clear explanations about academic integrity can reduce the number of cases of plagiarism [8].

To deal with opportunity, instructors need to avoid reusing assessments [9], encourage the use of student case studies [10], or impose additional grading methods such as oral presentation [11]. There is also a need to check student programs for plagiarism and penalize the perpetrators. Typically, instructors are aided with automated tools to identify similar programs. Some of the tools are MOSS [12], Sherlock [13], Plaggie [14], and SSTRANGE [15] (more publicly available automated tools can be seen in [16]). It is worth noting that instructors need to investigate similar programs and find sufficient evidence for plagiarism (burden of proof [13]). The high similarity is not always a result of plagiarism [17], and some similarities are coincidental [18].

Most of the tools mentioned in the previous paragraph are based on artificial intelligence (AI) which provides machine learning models to predict the occurrences of some events, including plagiarism. Beyond that, AI has also shown its possibilities in education such as supporting student engagements [19], personal tutoring system [20], information searching [21], and even generating exam questions [22]. Recently, another possibility is coming and interrupting the traditional ways of information gathering in the form of ChatGPT which is based on a large language model [23]. By using ChatGPT people can search beyond factoid-based information. Generated-based content such as story-telling and programming codes are ready to be presented by ChatGPT.

Misuse of ChatGPT introduces a new way of plagiarism in programming: students can ask the tool to write the solution (or at least part of it) at which the solution is generated from other people's work without citing the source(s), and thus lack of accountability [23]. Lecturers need to talk about the ethics of ChatGPT with their students. Due to the recency of ChatGPT misuse, to the best of our knowledge, there are no studies formally measuring the inappropriate benefits of using ChatGPT in programming education and reporting how to manually identify programs aided with ChatGPT.

In response to the aforementioned gap, we present a controlled experiment about inappropriate benefits and identification of ChatGPT misuse in programming education. The experiment involves 17 computing undergraduates who appear to know how their colleagues can breach academic integrity. These students are nominated based on academic values and attitudes that will guarantee their integrity and not misuse the tool. We also asked about their perspective on ChatGPT. Our study has thus the following research questions: 1) RQ1: How substantial are the inappropriate benefits of ChatGPT misuse in programming tests? 2) RQ2: What programming features can be used to identify ChatGPT-aided programs?

## 2 Method

A controlled experiment was conducted to address the research questions. RQ1 was addressed by comparing completion time and programming performance between tests with and without the help of ChatGPT. Students with ChatGPT appeared to have unfair benefits if they completed the test faster or they got higher marks in a statistically significant manner. RQ2 was addressed by asking students which programming features could be useful to identify ChatGPT-aided programs after they had used the tool for a test. We also asked several additional questions to understand more about the student perspective regarding ChatGPT.

Twenty computing undergraduate students were invited to participate with 17 of them accepting the invitation. They are never involved in plagiarism, and they have good impressions from instructors at our faculty. We are unable to consider more students as some of them might learn to misuse ChatGPT via our experiment and our current education environment is not fully ready to handle such a situation. While our selected computing undergraduates are never involved in plagiarism, they know how their colleagues do plagiarism as most of them are former tutors and/or are actively engaged with other students (as a member of the student senate or as laboratories staff). Each student was incentivized with Rp. 72.000 e-money (around 5 USD, sufficient to cover up to two-day meals for undergraduates)

The students were asked to complete the mid-test and final test from the previous offering of introductory programming (second semester of the 2021/2022 academic year). The course covers basic concepts of programming in Python: input, output, branching, looping, function, array, matrices, searching, and sorting. The first five concepts were covered in the mid-test while all of them were covered in the final test. Per the test, four programming tasks were introduced with 25 of 100 marks each. Details of the tasks in both tests can be seen in Table 1. The tasks were given in more detail with input-output examples. A test was expected to be completed in two hours. All students passed the course with A or B+ marks.

**Table 1.** Task details.

| ID | Task | Test | Difficulty |
|---|---|---|---|
| T1 | Money conversion | Mid | Easy |
| T2 | Statistics of students' GPAs | Mid | Medium |
| T3 | Shop invoice | Mid | Medium |
| T4 | 2D pattern of a combination of symbols and numbers | Mid | Challenging |
| T5 | Calculator | Final | Easy |
| T6 | Searching emoticons in a string | Final | Medium |

| ID | Task | Test | Difficulty |
|---|---|---|---|
| T7 | The closest and the farthest 2D coordinates from a particular coordinate | Final | Challenging |
| T8 | Advanced searching and sorting of items | Final | Challenging |

The students were split into two groups. The first group (nine students) completed the mid-test by themselves and the final test with the help of ChatGPT. The second group (eight students) completed the mid-test with ChatGPT and the final test by themselves. They were asked to record their completion time per programming task. In their corresponding ChatGPT session, they were also asked to report their step-by-step interactions with ChatGPT and which programming factors that could be useful for identifying ChatGPT-aided works. We also asked them to answer several additional questions to understand more about their perspective of ChatGPT. The complete set of survey questions can be seen in Table 2.

**Table 2.** Complete set of the survey questions.

| ID | Question |
|---|---|
| Q01 | Which test was done with the ChatGPT? |
| Q02-Q05 | Write down the time required to complete mid test task 1 to task 4 |
| Q06-Q09 | Write down the time required to complete final test task 1 to task 4 |
| Q10-Q13 | Elaborate how ChatGPT was used to complete task 1 to task 4 of given test (depend on the group the participant is in) |
| Q14 | ChatGPT can help students who are not proficient in programming to complete programming tasks |
| Q15 | ChatGPT-generated code can be easily understood by students who are not proficient in programming |
| Q16 | Students who are not proficient in programming are likely to use ChatGPT to complete programming tasks |
| Q17 | If the participant is a tutor, they are able to differentiate ChatGPT-generated code from original submissions |
| Q18 | Which factors that might be useful to differentiate ChatGPT-generated code from original submissions |
| Q19 | The participant believes with the correctness of ChatGPT-generated code |
| Q20 | Explain the reason for Q19 response |
| Q21 | The participant recommends the use of ChatGPT for completing programming tasks so long as it is properly cited in the comments |
| Q22 | Explain the reason for Q21 response |
| Q23 | The participant recommends the use of ChatGPT for completing general tasks so long as it is properly cited in the comments |
| Q24 | Explain the reason for Q23 response |

The first nine questions were introduced to address RQ1; Q01 was used to determine the group the participant is in; Q02-Q05 and Q06-Q09 were used to capture the completion time of each task. Question Q18 was introduced to address RQ2. The rest of the questions were introduced to understand the student perspective of ChatGPT. Questions Q14-Q17, Q19, Q21, and Q23 should be responded to on a 5-point Likert scale where 1 represents strongly disagree and 5 represents strongly agree. Questions Q10-Q13, Q18, Q20, Q22, and Q24 are open-ended questions.

RQ1 was addressed via completion time in seconds and programming performance, both per task and overall. Students with ChatGPT were compared with students without ChatGPT and they were favored with the use of the tool if their average completion time is shorter or their programming performance is better (i.e., higher mark) after being validated with an unpaired t-test with a 95% confidence rate. One of our authors is the instructor of introductory programming and they were in charge of marking the tests for measuring programming performance.

To identify the programming features that can be used to identify ChatGPT-aided programs in RQ2, some open-ended questions were asked to the participants. These questions are intended to extract the characteristics of how the students deal with the keywords when they use ChatGPT. We are interested to analyze how the core keywords, apart from the textual description in the programming assignments, would be explored by the students.

To achieve that objective, we propose a topic relation graph descriptive analyses which are based on the topic modeling [24]. The analysis consists of the following four main steps. In the first steps, the textual answers from each survey form are extracted. For each survey question, a bigram-based Latent Dirichlet Allocation (LDA) Topic Modeling is formed [25]. A bigram model is a sequence of two adjacent elements from a string of words that are expected to form relevant words with stronger semantic relations in the textual description.

The coherence scores are used to limit the number of effective topics, which are tuned for the number of topics from 2 to 10 with 10 words in each topic. The coherence scores measure how likely the words in a topic are semantically related. The higher the coherence score is, the more that the words in a topic are considered semantically related [26]. After the number of topics is determined, a topic relation graph is constructed to determine how one word is related to the other words in a topic and how they are related to other topics. The topic relation graph would also be useful to control the topic centrality. It would also be beneficial to extract core keywords for constructing hypothetical concepts for the open-ended questions in the surveys.

In short, the following steps are executed: first, all unique words in the optimized number of topics as nodes are extracted. Following that, the edges between the highest weighted word in a topic with the overlapping words in other topics are created. In this way, associated concepts within the window boundary, i.e., a path of associated keywords in the bigram model will be determined [27].

# 3 Result and Discussion

## 3.1 Inappropriate benefits of ChatGPT

Table 3 shows that in the mid-test, students with ChatGPT generally had shorter completion time (around half than that of students without ChatGPT) and the difference is statistically significant according to a two-tailed unpaired t-test with a 95% confidence rate ($p<0.001$). While observing per task, the difference is only statistically significant on T2 ($p<0,01$) and T3 ($p<0.01$). Calculating student GPAs and generating shop invoices are quite common and ChatGPT could help students well. T1 (money conversion) is also a common task but since it is quite easy to solve, the completion time is not significantly affected. In terms of programming performance, students with ChatGPT had comparable performance to those without ChatGPT. All differences are insignificant according to a two-tailed unpaired t-test with a 95% confidence rate.

**Table 3.** Averaged mid-test result.

| Task ID | Non-ChatGPT | | ChatGPT | |
|---|---|---|---|---|
| | Mark | Time (minutes) | Mark | Time (minutes) |
| T1 | 21.56 | 12.44 | 21.63 | 9.13 |
| T2 | 21.67 | 34.89 | 22.88 | 10.38 |
| T3 | 22.75 | 30.67 | 20.89 | 16.50 |
| T4 | 10.50 | 29.56 | 13.78 | 20.00 |
| T1-T4 | 77.75 | 107.56 | 77.80 | 56.01 |

For the final test, Table 4 depicts that the findings are somewhat similar to those of the mid-test. Students with ChatGPT completed the tasks two times faster than those without ChatGPT ($p<0.001$) but they had comparable programming performance. Further observation shows that differences in the completion time are also statistically significant for most tasks: T5 with $p<0.01$, T6 with $p<0.001$, and T8 with $p<0.01$. T7 is the only task at which students with ChatGPT had comparable completion times with their counterparts. Informal post-discussion with the students notes that for T7, ChatGPT-generated code had a slightly different purpose and needed further alignment to the task.

**Table 4.** Averaged final-test result.

| Task ID | Non-ChatGPT | | ChatGPT | |
|---|---|---|---|---|
| | Mark | Time (minutes) | Mark | Time (minutes) |
| T5 | 22.38 | 20.38 | 22.33 | 11.89 |
| T6 | 19.75 | 24.88 | 17.67 | 5.67 |
| T7 | 15.13 | 29.25 | 19.33 | 16.33 |
| T8 | 12.38 | 29.69 | 13.67 | 12.78 |
| T5-T8 | 69.63 | 104.19 | 73.00 | 46.67 |

### 3.2 Programming features to identify ChatGPT-aided programs

In the open-ended surveys Q10-Q13, the participants are requested to explain their experiences in detail to complete the programming assignments. Since we are interested in the core keywords that the participants used during the task completion in general, during the analyses we mixed all the answers of Q10-Q13. Another reason for this approach is that we hypothesize that each participant would have a particular specific strategy to interact with ChatGPT.

By performing the LDA topic modeling, the number of effective topics for Q10-Q13 is 8 with a coherence score of 0.749. The topic relation graph of these questions can be followed in Fig. 1. In this figure can be inferred that the participants are mainly using a copy-and-paste strategy from the textual description of the programming assignments. This can be seen in the core keywords as suggested by the topic relation graph. Some keywords, such as GPA, rupiah, rectangle, coordinate, box, and discount are initially coming from the textual descriptions of the assignments.

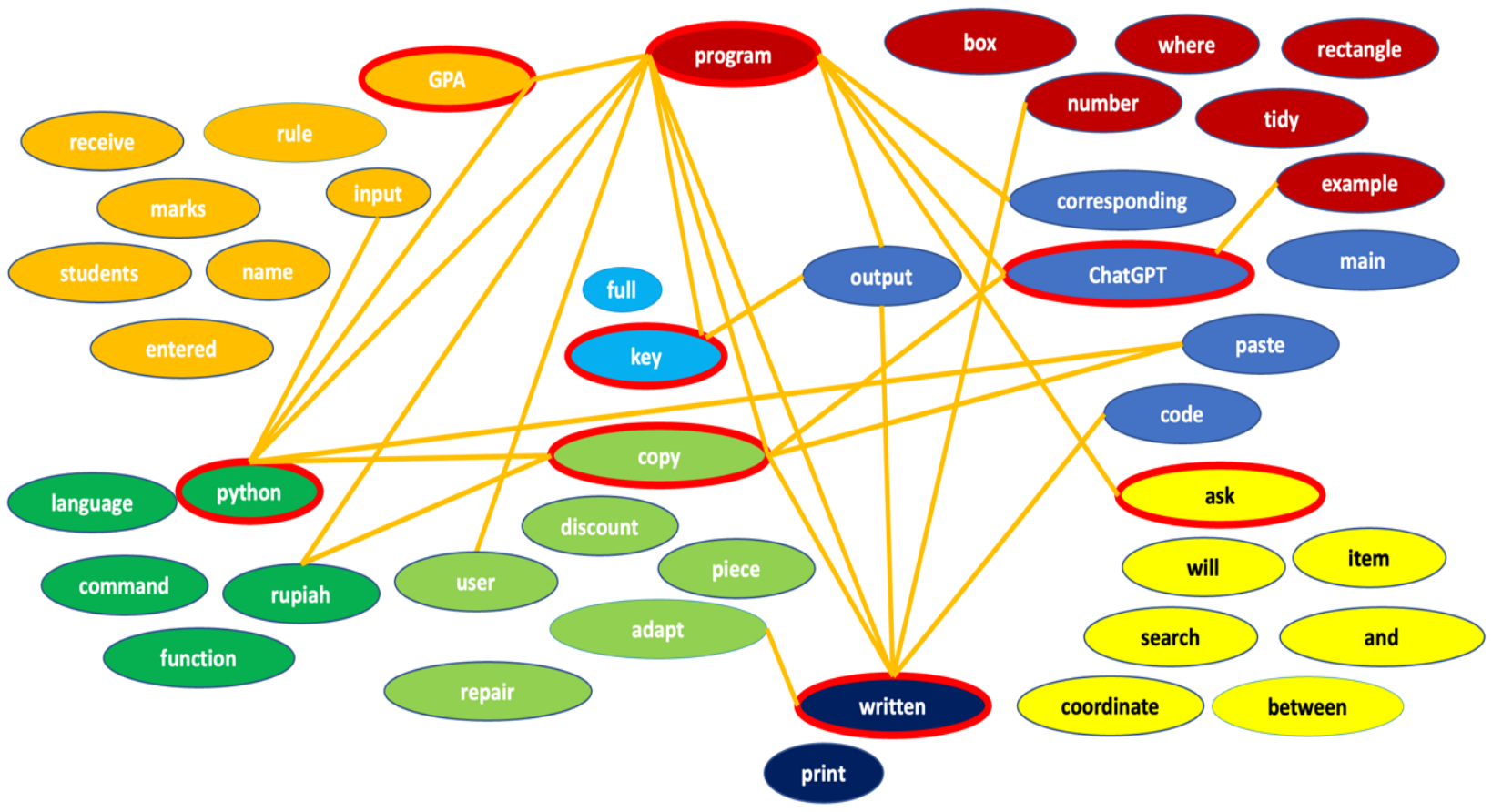

**Fig. 1.** Topic relation graph for Q10-Q13 (general strategy to use ChatGPT).

Several participants made special copy-and-paste strategies by using variations of the input and output formats in the program and also considering the programming language as described in the assignments. After they found the suitable retrieved code example from ChatGPT, in the subsequent steps they will try to adapt the code and repair it as completely as possible according to the assignment. Some also try to translate the assignment into English and enter it into ChatGPT. They made subsequently some adjustments to functions and variable naming.

It is interesting that although the participants have in general almost the same copy-and-paste strategy, they have the confidence to differentiate the ChatGPT-generated codes and those which were written by relatively new programmers. The respondents were confirming that ChatGPT will be very helpful for students who have programming skills limitations. Around 41% of the respondents strongly agree, and around 35% agree about this fact. But, on the other hand, there are some critical characteristics that ChatGPT has, in comparison to the standard coding style. This is confirmed by the results of Q17, which state that around 30% of the respondents strongly agree that they can differentiate codes generated by ChatGPT, and around 35% agree about the fact.

In further analysis of Q18, we explored which factors that might be useful to differentiate ChatGPT-generated code from original submissions based on the participants' point-of-view. The topic relation graph of Q18 can be seen in Fig. 2, which consists of 7 topics and a coherence score of 0.692. The participants describe that computer codes generated by ChatGPT have structural and syntactical advances in comparison to those written by beginners.

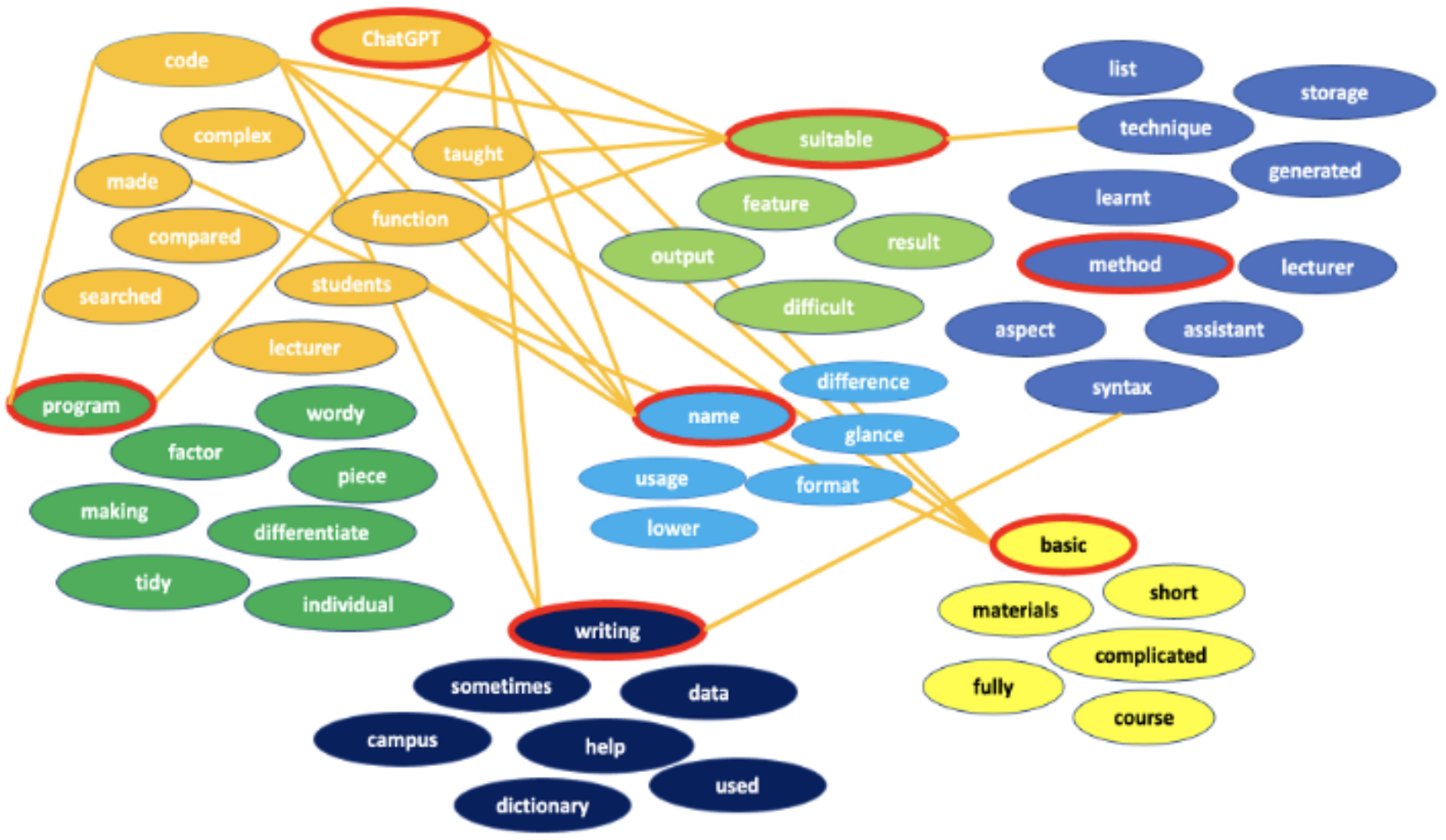

**Fig. 2.** Topic relation graph for Q18 (factors to differentiate ChatGPT).

Some aspects, such as the strategy to efficiently usage of memory (storage) and function arrangement are rather advanced. This is by using, for instance, a list or a dictionary data structure. Further, the generated ChatGPT code itself is mostly tidier and shorter (efficient), than those developed by students. In some cases, it is obvious to observe that the generated codes by ChatGPT are beyond the typical course materials for first-year students.

### 3.3 Student perspective of ChatGPT

Further analyses from the student perspective regarding the use of ChatGPT are also important as a means to identify the tendency of ChatGPT usage during course assignments. In Q15, Q16, and Q19, we explored the general opinion of the participants on whether ChatGPT would be beneficial for them. In this sense, it is not only that they could answer the assignment correctly, but also emphasize ideas about the knowledge-containing material in the course. Commonly, a teaching assistant or a lecturer will provide some activities to confirm the course learning objectives, such as presentations or small quizzes.

From the answers to Q15, most of the participants (47%) do not agree and strongly do not agree (18%) that generated code of ChatGPT will be easily understood by newbies or those who are not proficient in programming. However, as teaching assistants, they have majority agreements that most non-proficient students would use ChatGPT-generated codes to fulfill their assignments (Q16), i.e., 29% strongly agree and 41% agree. These facts suggest that reconfirmation of submitted assignments would be important to guarantee the authenticity of the codes and to ensure the achievement of learning objectives as well. These facts are also supported by the results of Q19. The participants have a high appreciation for the generated codes by ChatGPT, and they consider the correctness of the generated codes (6% strongly agree, 35% agree, and 41% neutral) with some arguable opinions.

In the open-question survey of Q20, the participants were asked to express their opinion about their belief in the correctness of ChatGPT-generated code. The topic relation graph can be followed in Fig. 3. Four topics are considered effective with a coherence score of 0.654. We can infer that the participants have critical opinions regarding ChatGPT. To be considered correct, the code from ChatGPT should fulfill the expected output and results as described in the assignments. Further, the generated code needs to be written properly as taught in the course, such as how the classes and functions are defined with acceptable factoring and suitable data structures.

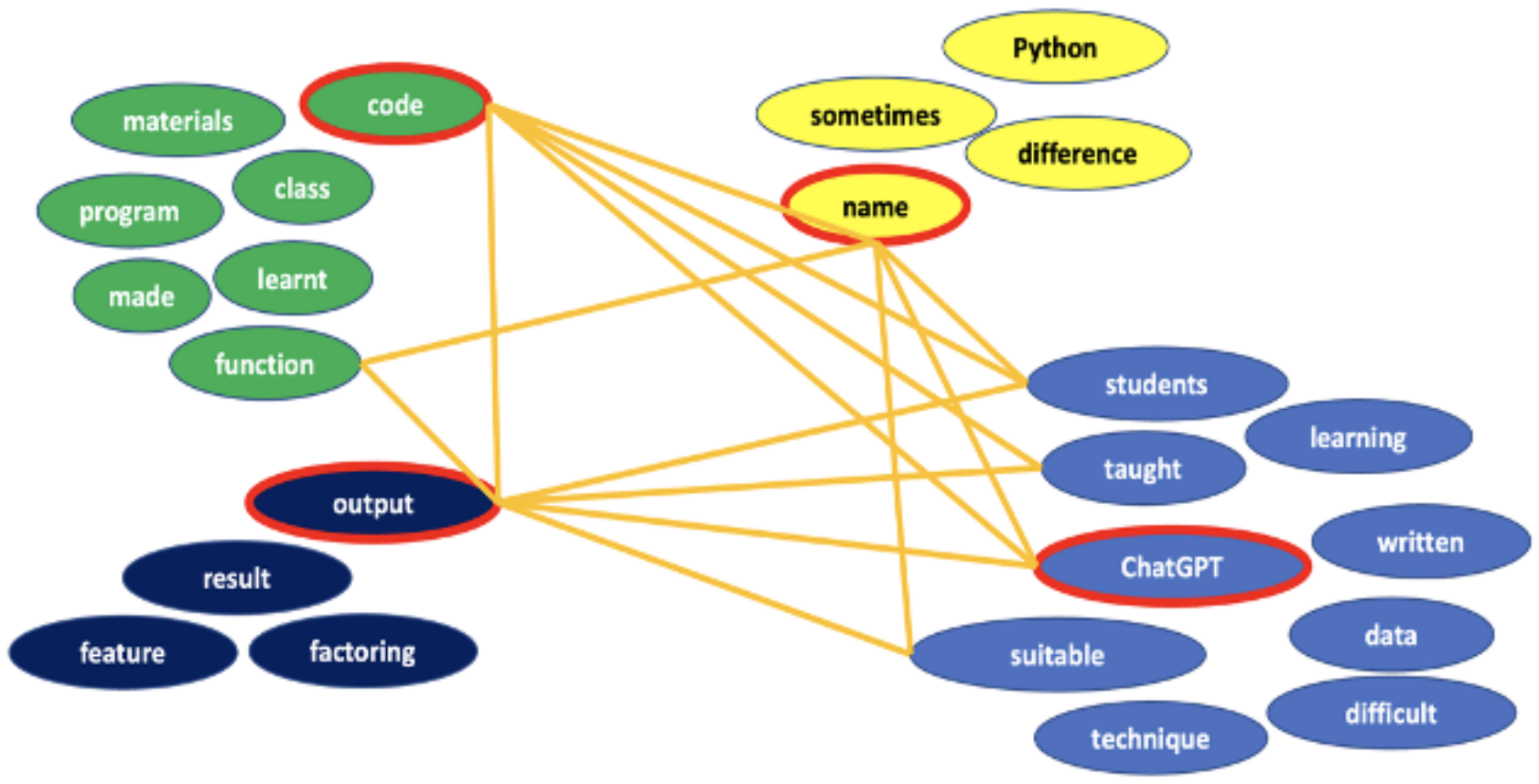

**Fig. 3.** Topic relation graph for Q20 (participants' beliefs in the correctness of ChatGPT).

Some questions in Q21 and Q23 are being asked to investigate the participants' recommendation of using ChatGPT in programming and general tasks. Almost half of the participants strongly agree (12%) and agree (35%) to recommend the utilization of ChatGPT in programming tasks (Q21). In this case, it would be necessary to refer credits and citations to ChatGPT, although only part of the code uses the generated form. A comparable result can also be seen for Q23 which states that the majority of the respondents strongly agree (12%) and agree (47%) to use ChatGPT in general tasks beyond programming tasks.

Specific reasons for using ChatGPT in programming tasks are asked in Q22, and analyzed in a topic relation graph, with 3 effective topics and a coherence score of 0.612. ChatGPT would be useful as a means to help with programming tasks. Especially for first-year students, ChatGPT will be beneficial to help students understand the code and learn to program from code examples retrieved by ChatGPT. Thus, ChatGPT can be considered a web search engine such as Stack Overflow but gives more information and directly focuses the result on highly probable right answers. Nevertheless, ChatGPT can be used to validate computer program codes, and this will be useful for students to learn how to write an acceptable computer program. Comparable comments are also applied for general assignment tasks beyond programming in Q24. Our survey participants stressed the important aspect of logical and critical thinking when using ChatGPT. ChatGPT is indeed a trend, but it must be used wisely, especially in academics.

## 4 Conclusion and Future Work

In this research, we have identified ChatGPT misuse in programming assignments for first-year computer science students in a controlled experimental environment. The main strategy when using ChatGPT in programming tasks as suggested by our respondents is by copy-and-paste approach. The most used part to be copied to ChatGPT is the task descriptions, followed by the input and output structures.

The retrieved and generated code by ChatGPT is subsequently adapted to variable and function naming. Although it seems easy to retrieve the code, the ChatGPT codes are sometimes rather difficult to understand by those who just learn to program. The generated code is highly efficient and uses complex data structures like lists and dictionaries. Based on the survey results, ChatGPT is recommended to be used as an assistant to complete programming tasks and other general assignments. ChatGPT will be beneficial as a reference as other search engines do. Logical and critical thinking are needed to validate the result presented by ChatGPT.

For further research, we plan to identify the differences in the semantics and code structures of ChatGPT and human-generated code. It will be important to examine the control flow graphs of the codes. And thus, the ChatGPT will be beneficial as a means to extract important features to be (machine)-learned to code efficiently, and at the same time to avoid plagiarism.

### Acknowledgment

The research presented in this paper was supported by the Research Institute and Community Service (LPPM) at Maranatha Christian University, Bandung, Indonesia.